\documentclass[letterpapper, 10pt, conference]{ieeeconf} 
\IEEEoverridecommandlockouts
\usepackage{amsmath,amssymb,amsfonts}
\usepackage[utf8]{inputenc}
\usepackage[english]{babel}
\usepackage{algorithmic}
\usepackage{graphicx}
\usepackage{subcaption}
\usepackage{textcomp}
\usepackage{xcolor, colortbl}
\usepackage{csquotes}
\usepackage{svg}
\usepackage[export]{adjustbox}
\usepackage{fixltx2e}
\usepackage{subcaption}
\usepackage{bm}
\usepackage{todonotes}
\usepackage{dblfloatfix}
\usepackage{hyperref}
\hypersetup{
    colorlinks=true,
    linkcolor=black,
    filecolor=magenta,      
    urlcolor=cyan,
    citecolor=black,
    pdftitle={Teleoperation},
    }

\newcommand{\bx}{{\mathbf{x}}}

\newcommand{\bR}{{\mathbf{R}}}

\newcommand{\bT}{{\mathbf{T}}}

\newcommand{\fra}[1]{{\mathcal{F}_{#1}}}
\newcommand{\fref}[1]{Fig.~\ref{fig:#1}}

\definecolor{blueTransparent}{rgb}{0.8,0.8,1.0}
\definecolor{redTransparent}{rgb}{1.0,0.8,0.8}
\definecolor{purpleTransparent}{rgb}{1.0,0.5,1.0}

\begin{document}
\title{Teleoperation of a robotic manipulator in peri-personal space:\\ a virtual wand approach}

\author{Alexis Poignant$^{1}$, Guillaume Morel$^{1}$, Nathanaël Jarrassé$^{1,2}$
\thanks{$^{1}$Sorbonne Université, CNRS, INSERM, Institute for Intelligent Systems and Robotics (ISIR), Paris, France.}%
\thanks{$^{2}$Email: jarrasse@isir.upmc.fr}%
}

\maketitle

\begin{abstract}
The paper deals with the well-known problem of teleoperating a robotic arm along six degrees of freedom. The prevailing and most effective approach to this problem involves a direct position-to-position mapping, imposing robotic end-effector movements that mirrors those of the user, \fref{mapping}-top. In the particular case where the robot stands near the operator, there are alternatives to this approach. Drawing inspiration from head pointers utilized in the 1980s, originally designed to enable drawing with limited head motions for tetraplegic individuals, we propose a "virtual wand" mapping. It employs a virtual rigid linkage between the hand and the robot's end-effector,  \fref{mapping}-bottom. With this approach, rotations produce amplified translations through a lever arm, creating a "rotation-to-position" coupling. This approach expands the translation workspace at the expense of a reduced rotation space. 

We compare the virtual wand approach to the one-to-one position mapping through the realization of 6-DoF reaching tasks. Results indicate that the two different mappings perform comparably well, are equally well-received by users, and exhibit similar motor control behaviors. Nevertheless, the virtual wand mapping is anticipated to outperform in tasks characterized by large translations and minimal effector rotations, whereas direct mapping is expected to demonstrate advantages in large rotations with minimal translations.
These results pave the way for new interactions and interfaces, particularly in disability assistance utilizing head movements (instead of hands). Leveraging body parts with substantial rotations could enable the accomplishment of tasks previously deemed infeasible with standard direct coupling interfaces.

\end{abstract}


\section{Introduction}
Over the past decade, the democratisation of 6 and 7 Degrees of Freedom (DoF) robotic manipulators has significantly increased their use in collaborative industrial and assistive applications \cite{noauthor_world_2023}. Despite these developments, controlling 6 DoF displacements (3 translations and 3 rotations) remains a challenge due to the need of 6-DoF interfaces, especially for position-position control. This paper concentrates on addressing the control issues associated with such robots in close proximity. This specific operational scenario is commonly encountered in enclosed-robot environments and for tasks requiring assistive functionality. For example, assisting individuals with restricted hand mobility \cite{chung_functional_2013} \cite{al-halimi_performing_2016} \cite{casadio_body-machine_2012}, or industrial operators with occupied hands, such as in 3-hands soldering of large parts \cite{parietti_supernumerary_nodate}. It can also be found in environments with nuclear or chemical hazards, which may require manipulation through a glove-box \cite{pulgarin_assessing_nodate}.

\begin{figure}[t!]
    \centering
    \includegraphics[width=0.75\linewidth]{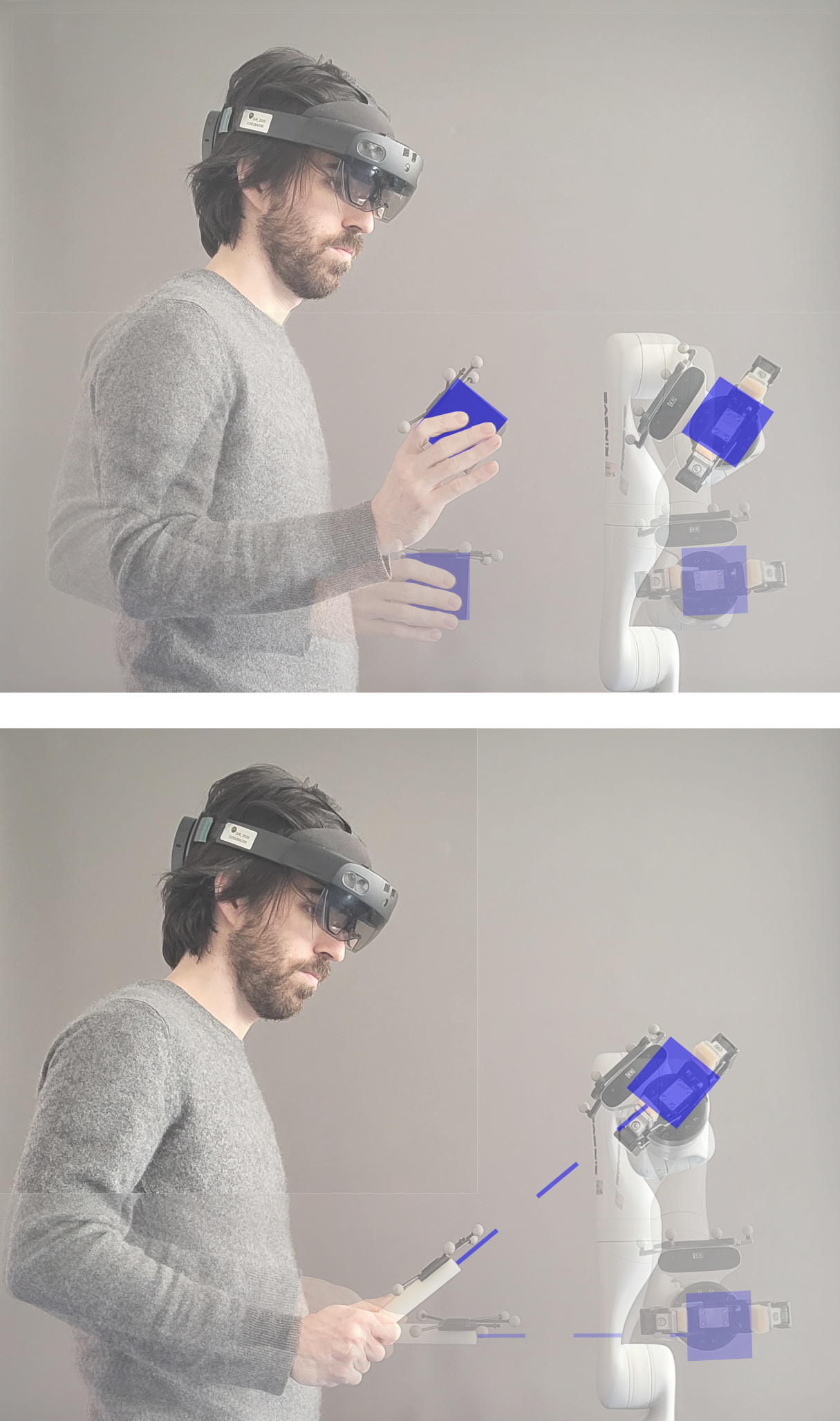}
    \caption{A comparison of the one-to-one direct mapping mapping (top) and wand (bottom) between a start (lighter) and an end (darker) configuration. In the wand mapping, rotations of the hand produces amplified translations coupled with rotations of the end-effector, while one-to-one direct mapping decouples translations and rotations of the end-effector.}
    \label{fig:mapping}
\end{figure}

In such scenario, the operational distance being short, it allows for direct visual feedback from the robot, as opposed to relying solely on a screen. This not only provides a clearer view of the environment but also diminishes the reliance on artificial haptic feedback and reduces the delay between the robot's actions and the user's responses. It also increases the potential feelings of embodiment of the robot for the user, as the proximity of the robot increases visual but also sound or proprioceptive-based feedback. 

Existing literature emphasizes that, in such scenarios, position-to-position control tends to yield superior performance and user preference over position-to-velocity control \cite{lincoln_visual_1953, won_kim_comparison_1987} such as joysticks. However, implementing a position-position control is challenging for two primary reasons. First, there is a deficiency of position-to-position interfaces, which are often large \cite{noauthor_virtuose_nodate}, and second, the direct one-to-one mapping of a user's body part to the robotic end-effector is restricted by the user's range of motion. For instance, if the robot's end-effector displaces by one meter, the user's corresponding body part would also have to move one meter, limiting practicality. Conversely, direct mapping proves highly effective for rotational displacements around a fixed point. To overcome the translational limitations in the task space, position-control interfaces often incorporate clutching mechanisms \cite{dominjon_comparison_nodate} (such as on Intuitive Surgical Da Vinci Robot \cite{freschi_technical_2013}). This allows users to temporarily disengage from the robot, return to their initial position, and then re-engage with the manipulator. Alternatively, scaling can be employed \cite{dominjon_comparison_nodate}, but this approach usually sacrifices precision and noise-sensitivity. However, addressing these challenges remains needed to enhance the usability and effectiveness of position controlled 6 DoF robotic manipulators.

In this paper, we propose an alternative mapping approach, specifically employing a lever, or a fixed-length "wand", to transform body rotations into amplified translations, and body translations into diminished rotations. In this mapping, the user is equipped with a rigid wand, and the virtual tip of the latter serves as an indicator for the desired position and orientation of the robotic manipulator. This mapping offers the advantage of expanding the translation workspace through the leverage provided by the lever and the rotations of the body. However, it comes at the cost of reducing the rotational workspace, necessitating the rotation of the user's hand around the robot's end-effector, and therefore large translations proportional to the wand's length.

By converting human rotational motions into translations, this mapping strategy also enables users to leverage body parts with limited translational range but significant rotational capability, such as the trunk or the head. The use of head-based wands was previously explored in the eighties and nineties, particularly with tetraplegic patients possessing residual head mobility \cite{jewell_custom-made_1989}\cite{dymond_controlling_1996}, employing head-based wands to draw or operate a keyboard. Wands are also known to be able to integrate into the body schema \cite{maravita_tools_2004}, making them intuitive. Nevertheless, traditional rigid wands have inherent limitations, including constraints on weight and size, as well as a lack of motorization and easy reconfiguration.

To overcome these limitations, we propose the use of a virtual wand instead of a physical one. A virtual wand eliminates these physical constraints, and offers more flexibility. However, as the robot motions are often slower than the human operator's motions, the robot effector might be delayed compared to the virtual lever's effector. Users can interact with the virtual wand without seeing it directly, but, in order to visualize the delay, they can opt to use an Augmented Reality (AR) headset to observe the virtual wand in real time. While the latter may introduce some visual discomfort due to the holograms, it provides an effective means to mitigate the impact of delays (which are usually undesirable for teleoperation \cite{won_kim_comparison_1987}), and should ease the integration of the tool into the body schema, enhancing the overall user experience. The use of Augmented Reality for distant teleoperation was previously explored in a one-to-one direct mapping scenario \cite{nuzzi_hands-free_2020}, but was limited to 2 DoF and not compared to other methods or control without AR visual-based feedback.

Our proposed Wand Mapping aims to provide an amplified position mapping that does not require any customization (unlike scaling) and is solely based on geometric relations, like one-to-one Direct position-to-position mapping. As it amplifies the operator's rotation into translation, we may call it "rotation-to-position" mapping. 

Our goal is not to replace the traditional one-to-one position-to-position mapping, but to propose an alternative that naturally amplifies body rotations, and therefore could be use through head or trunk motions, and therefore in the previously described scenarios, unlike position-to-position. In the following experimental results, we decided to use the operator's hand to control the robot, as it is the only body part that can naturally use both mappings with ease, and thus provide an equal comparison, but applied scenarios would a priori include assistance tasks, using the head or torso, and users with limited mobility.

In the following sections, we first describe the implementation of the system and 6 DoF mappings, and we then evaluate the performances, preferences and impacts on motor control of the Wand Mapping compared to the traditional and well-known one-to-one Direct Mapping.

\section{Material and Methods}
\subsection{Experimental set-up}
The set-up is presented in Fig.~\ref{fig:AR}. 
The virtual device base frame $\fra{E_H}$ is materialized by an optical marker attached to the tool manipulated by the user's hand, either a wand or a cube. The marker is tracked in real time by an Optitrack motion capture system w.r.t. its fixed frame $\fra{O}$. The experiment also involves a Hololens 2 Augmented Reality (AR) headset that is self-localized in real-time w.r.t. a fixed $\fra{W_{AR}}$, and a 7 DOFs \copyright Kinova GEN3 Ultra Light Weight robot with fixed base frame $\fra{B_R}$.

Prior to the experiment, a procedure allows for registering $\fra{O}$, $\fra{W_{AR}}$ and $\fra{B_R}$, all with respect to a fixed world frame $\fra{W}$ (chosen arbitrarily). This allows to express in the same frame $\fra{W}$ the optical measurements, the location of the pointer tip to be displayed in the AR headset, and the robot position or velocity commands.

The AR headset is used to display the desired end-effector location to the participant (in dark blue as seen Fig.~\ref{fig:AR}). This end-effector is a cube centered at the origin of frame $\fra{E_R^\star}$. Namely, the desired pose for the robot writes:
\begin{equation}
\bT_{W \to E_R^\star(t)} = 
    \begin{pmatrix}
    \bR_{W \to E_R^\star(t)} & \bx_{W \to E_R^\star(t)}\\
    0~~~~0~~~~0 & 1
    \end{pmatrix}~,
    \label{eq:desired_position_to_desired_pose}
\end{equation}
where $\bR_{A \to B}$ and $\bx_{A \to B}$ are the rotation matrix and the origin translation from $\fra{A}$ to $\fra{B}$, respectively.

The robot end-effector location $\bT_{W \to E_R}$ is servoed to $\bT_{W \to E_R^\star}(t)$ using a resolved rate controller that imposes an end-effector velocity proportional to the error. The controller guarantees, due to its integral effect, a null permanent error. The convergence rate is tuned by a simple proportional gain $k$ determining the closed-loop position control bandwidth. Its value is set to $k=0.5~s^{-1}$, heuristically determined to obtain the fastest behavior before exciting the internal joint position controller oscillations of the Kinova. The augmented reality (AR) headset additionally presents the actual position of the robot effector in a semi-transparent manner. This feature enables participants to visually estimate when the robot has attained the desired configuration.

An illustrative video detailing the system used with the head and torso can be seen at:  \url{https://www.youtube.com/watch?v=EgwzT784Fws}.

\subsection{Virtual Representation}

\begin{figure}[t!]
    \centering
    \includegraphics[width=0.75\linewidth]{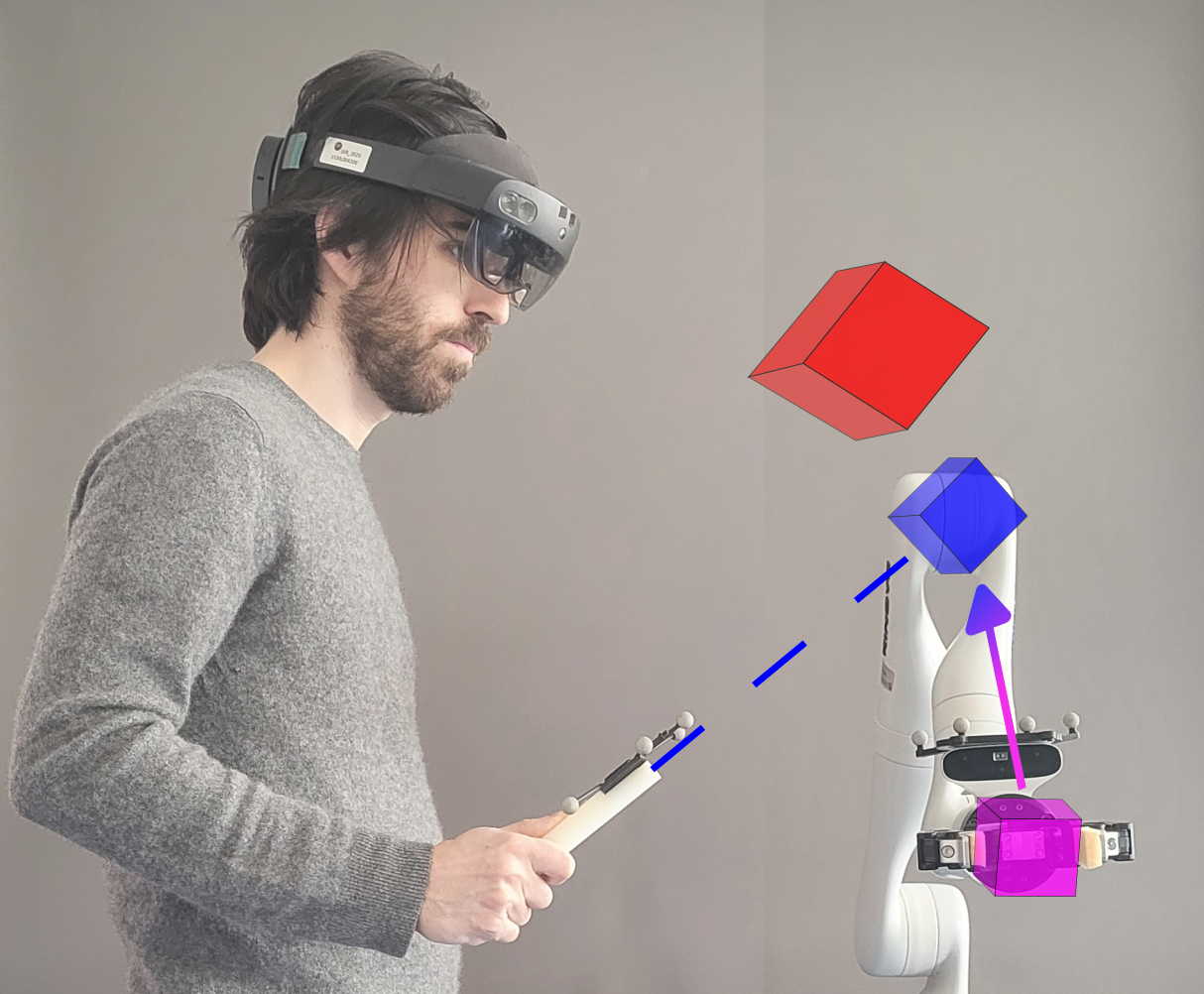}
    \caption{A representation of the AR visualisation during Wand Mode : the desired end-effector position is shown in blue (and does not appear during the visual perturbation mode), the target is shown in red, and the current robot's end effector is shown in magenta. The robot is servoed to follow the desired configuration (arrow, not shown in the AR headset). A similar representation is show in Direct mapping, without the handle of the hand and solely the desired blue end effector.}
    \label{fig:AR}
\end{figure}

The virtual environment features three objects: a light magenta cube representing the current configuration of the robot end-effector, a dark blue cube denoting the desired end-effector configuration (either corresponding to the tip of the wand or the teleoperated effector which directly maps the hand motions), and a red cube signifying a target location that the robot end-effector is expected to reach such as seen Fig.\ref{fig:AR}.

\subsection{Teleoperation control modes}
\subsubsection{Direct Mapping}
The direct mapping consists of simply mapping the displacement of the hand between the initial start of the experiment $t_0$ and the current instant $t$ to the desired robot effector:
\begin{equation}
    \bT_{E_R^\star(t_0) \to E_R^\star(t)} = \bT_{E_H(t_0) \to E_H(t)}
\end{equation}
If at the initial time both the hand and the robotic effector have a similar orientation, then the user has the illusion that the desired effector copies its motions with a constant translation in-between. To reinforce the illusion, users are holding a real cube of the dimensions of the virtual one.

\subsubsection{Wand Mapping}
The wand mapping consists of giving the illusion that the user is holding a wand whose tip is the desired cubic end-effector of the robot. The mapping writes:
\begin{equation}
    \bT_{W \to E_R^\star(t)} = \bT_{W \to E_H(t)}\bT_{E_H(t) \to E_R^\star(t)}
\end{equation}
where $\bT_{E_H(t) \to E_R^\star(t)}$ is constant and equal to its initial configuration, which corresponds to the geometry of the virtual wand. To give the illusion of a real wand, participants are holding a stick that they point towards the robot end effector at $t_0$.

\subsubsection{Integration of a visual perturbation}
In order to evaluate the integration of the tool and of the mapping, we also propose at specific times to delete the visualization of the desired blue effector in direct mapping, and of blue the wand in the wand mapping. Users still hold the real object (cube or stick) and still see the half-transparent current position of the robot (magenta). The behaviour of the robot is exactly the same, meaning that, if the user stop their motion, the robot end-effector's will attain the invisible desired end-effector configuration. This mode is equivalent to introducing a delay in the command, which was visually resolved using the AR. It is used in the experimental section to study the integration of the mappings in the body schema.

\subsection{Protocol}
Prior to commencing the experimental session, participants were introduced to the Augmented Reality Environment. Instructions were provided regarding the mappings, encouraging participants to "focus on the task" and telling them that they could "move without any restrictions," allowing walking if necessary, thereby granting maximum freedom to the participants. Each participant used both control modes. The experiment comprised 7 trials for each control mode. Half of the participants initiated the experiment using the Direct Mapping control mode, while the remaining half began with the Wand Mapping control mode. 

For each operational mode, the 4th and 6th trials were executed without the blue visualization, whereas the remaining trials featured the complete augmented reality (AR) visualization. Subjects were given the liberty to rest for as long as they desired between trials. At the onset of each trial, the participants' hands were consistently positioned at the same starting point, indicated in the AR headset. This starting point corresponded to a 45cm distance (and a 45cm wand in Wand Mode) between the desired effector and the center of the participant's palm.

Each of the 7 trials comprised the task of reaching a total of 30 targets — 15 central points and 15 outer targets. Targets alternated between the central point and outer locations, creating 15 back-and-forth trajectories. For outer targets, a 15cm hand translation (determined using a Fibonacci distribution on a 3D sphere \cite{gonzalez_measurement_2010} in order to explore various directions) was required, as well as a rotation between 0° and 45° around a randomly uniformly chosen 3D-axis. The order of the targets was consistent between trials, which participants were informed.

The \textbf{targets were not the same in both modes, but required the same hand starting and end points}, which allowed to have consistent and comparable performances and hand trajectories between the modes. Certainly, employing identical targets for both mappings would inherently bias one method over the other. The one-to-one mapping is inherently proficient for tasks involving substantial rotations without translation, while the wand mapping is more efficient to manage extensive circular translations. Selecting uniform targets would have introduced disparities in mapping efficiency, and optimal hand paths with varying lengths, leading to incomparable trajectories and task completion times. Hence, our emphasis was on ensuring uniform optimal hand motions between both mappings.

The set of targets also remained the same across trials, which participants were duly informed. Successful target reaching was acknowledged when the robot remained within the 2cm tolerance for at least 1 second, with a 10-degree tolerance in axis-angle representation relative to the target (represented by a large red cube, which turned green when the configuration of the robot was correct). Once achieved, the target disappeared, and the next one appeared. The chosen tolerances struck a balance between requiring precision from the participants and ensuring a rapidly feasible task. Given the potential challenges associated with depth perception and precise orientation in AR visualization, the chosen tolerances were designed to assess the intuitiveness of the mapping rather than focusing on participants' perception of AR objects.

The total completion time for each set of 7 trials in each mode ranged from 25 to 30 minutes. The overall duration of the experiment, encompassing instructions and rest sessions, fell between 1 hour and 1 hour and 15 minutes.

\subsection{Questionnaires}
Additionally, after the whole experimental session, participants were asked to fill questionnaires with six questions. From French to English they can be translated as:
\begin{enumerate}
    \item To what extent has the task required a physical effort?
\item To what extent has the task required a cognitive effort?
\item How much did you feel in the control of the robotic arm?
\item How intuitive was it to control the robotic arm?
\item To what extent have you felt that the robot was an extension of yourself?
\item Which global score would you rate this experiment?
\end{enumerate}

Participants answered with a score ranging from -3 (very little) to 3 (very much).

One questionnaire was filled for each control mode, with and without the visualization of the desired effector / of the wand, resulting in a total of 4 questionnaires. For the reading of the results in the next section it is worth remembering that a high rated answer for questions~1 and~2 corresponds to a negative characteristic of the system (high load), while, for questions~3 to~6 it reflects a positive characteristic (intuitiveness, general satisfaction, etc.).

Questionnaires groups are compared using a paired Mann-Whitney Wilcoxon U-Test as data are non-normal according to the Shapiro-Wilks Test.

\subsection{Participants}
The experimental study was carried out in accordance with the recommendations of Sorbonne Universit\'e ethics committee CER-SU, which approved the protocol. 

Twenty asymptomatic participants, aged 18-55, volunteered for this experimental study. 
They all gave their informed consent, in accordance with the Declaration of Helsinki.

\section{Results}
The success rate was of 100\% for all participants and all targets in both modes, with and without visualisation. Distributions presented in this section show the median, the 25th and 75th quartile, as well as 1.96 time the standard deviation in dotted line, which correspond to approximately the 95\% confidence interval. Rotations are denoted using the scalar geodesic angle from the angle-axis representation.

\subsection{Performances of the effector}
\subsubsection{Task Performances}
\begin{figure}[ht!]
    \centering
    \includegraphics[width=0.75\linewidth]{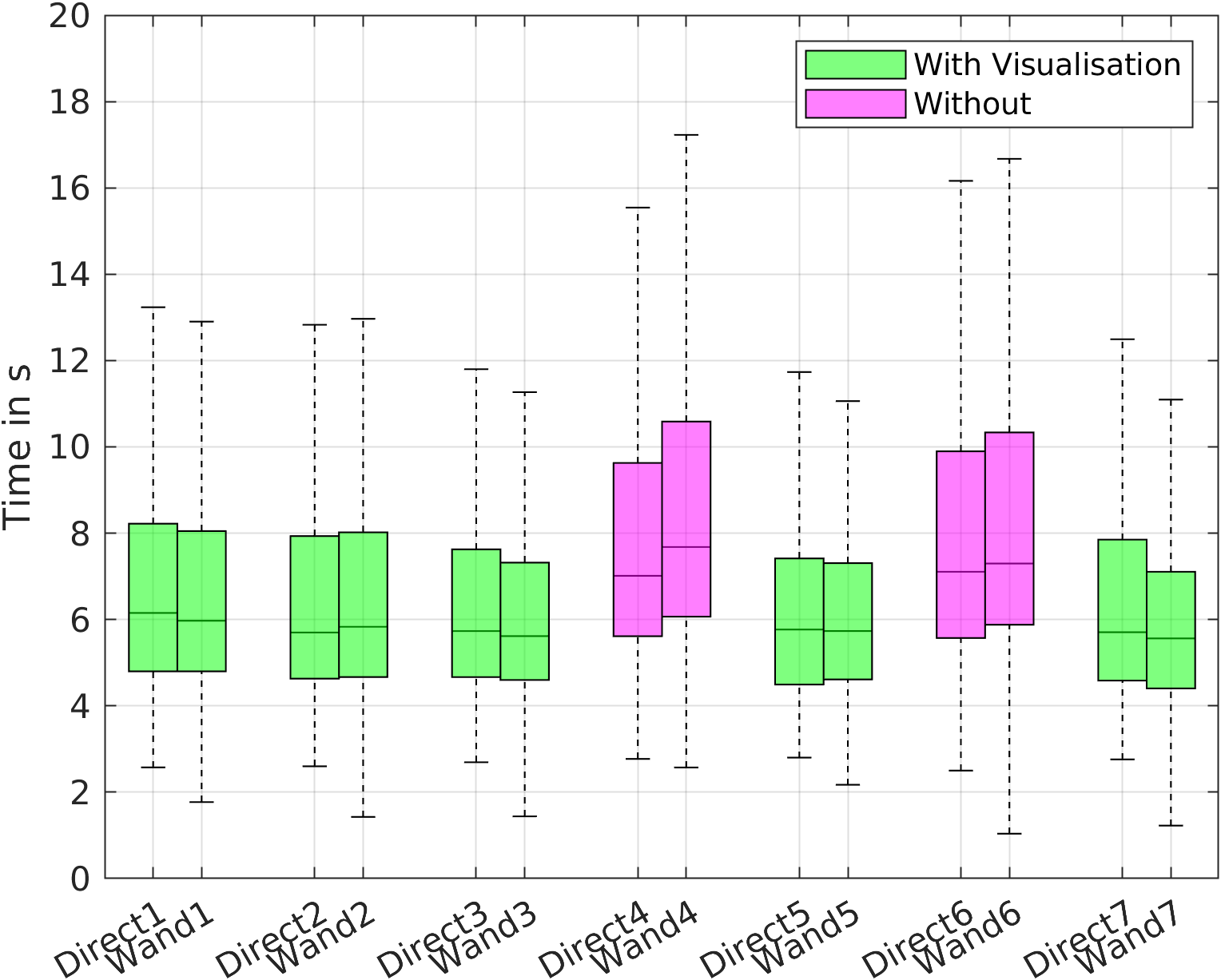}
    \caption{Distribution of the duration per target in s. Green denotes a trial with visualization, and magenta without.}
    \label{fig:times}
\end{figure}

We show on Fig.\ref{fig:times} the distribution of task durations to complete one target, per trial, and per mode of mapping across all participants. The disappearance of the visualisation shows an increase over time in both modes. Otherwise, in all trials with visualisation, the median time per target remains relatively constant, as well as the quartiles.

\subsubsection{Overshoot}
\begin{figure}[ht!]
    \centering
    \includegraphics[width=0.75\linewidth]{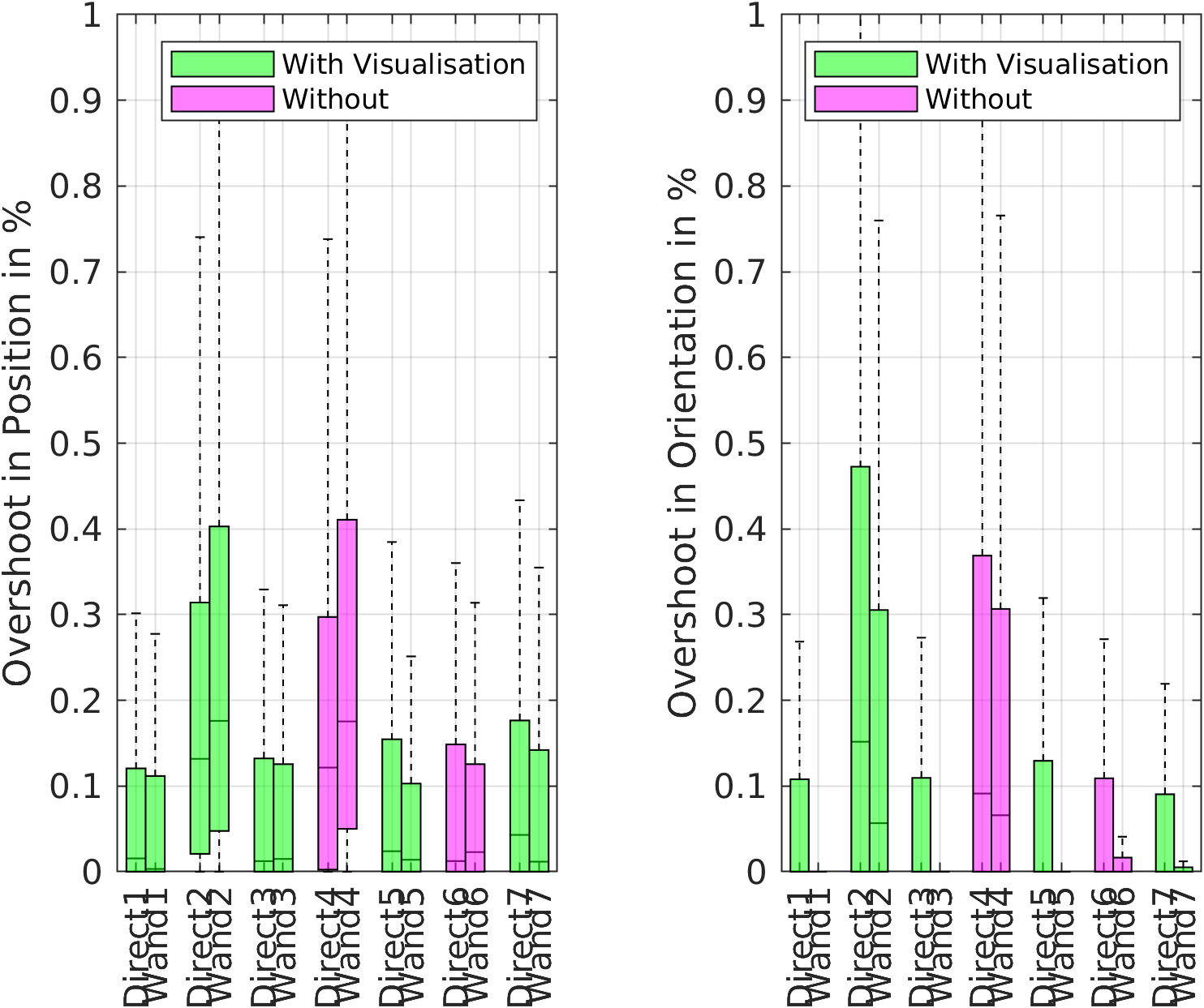}
    \caption{Distribution of the overshoot of the desired virtual end-effector above the tolerance per target as a percentage, both in translation and in rotation. Green denotes a trial with visualization, and magenta without.}
    \label{fig:overshoots}
\end{figure}

In order to evaluate the precision, we also show on Fig.\ref{fig:overshoots} the distribution of overshoots above the target tolerance as a percentage, both in translation and rotation. The first visualisation removal (4th trial) shows an increase of the overshoot, but this phenomenon is mitigated during the second trial without visualisation (6th trial). We also observe a strong overshoot during the 2nd trial for both modes.

\subsubsection{End effector Motion}
\begin{figure}[ht!]
    \centering
    \includegraphics[width=0.75\linewidth]{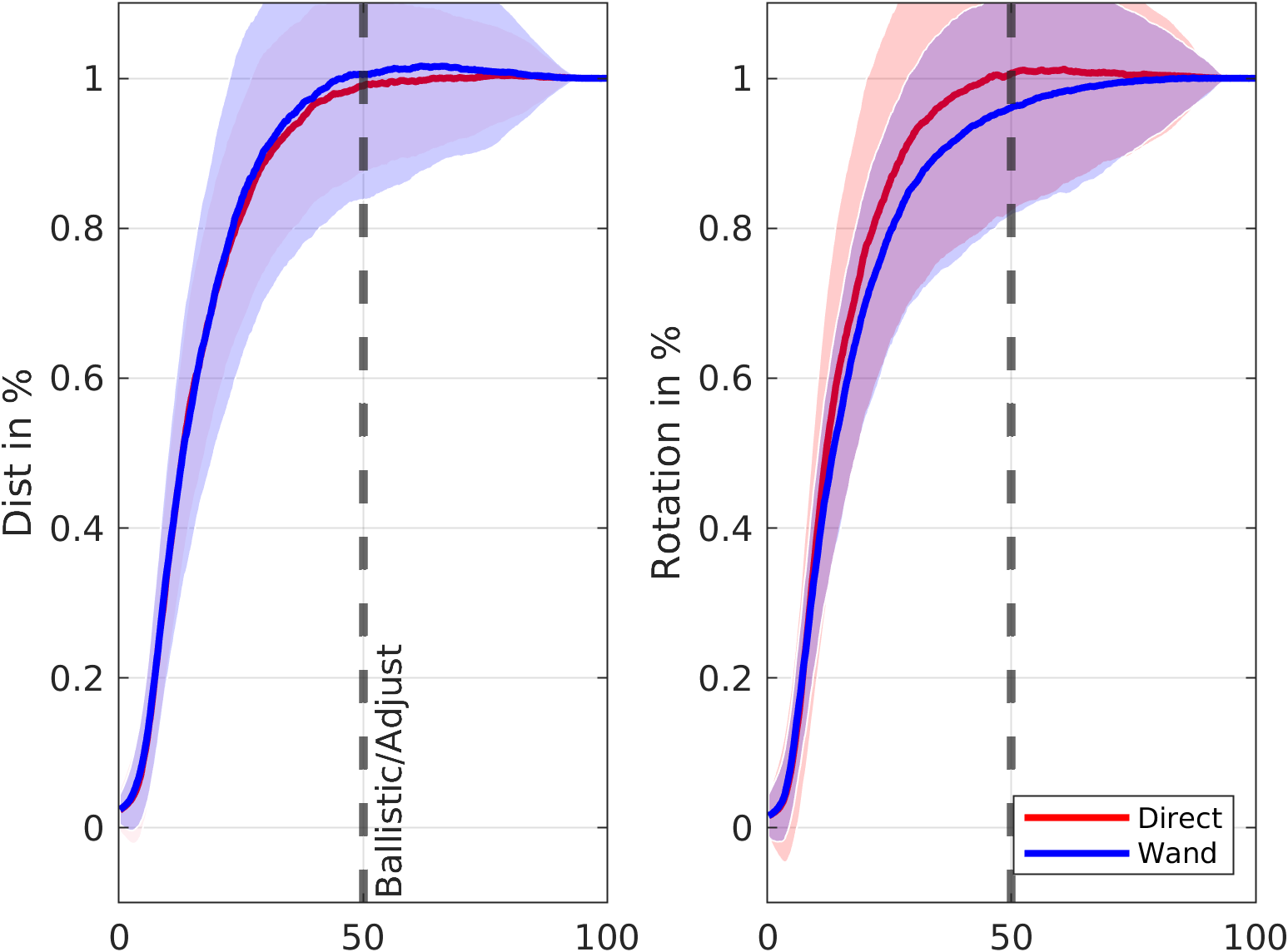}
    \caption{Normalized Median of the Hand Motion in Translation and Rotation during trials with Visualisation. The lighter area denotes the standard deviation of the curve. The dotted line indicates the separation between the estimated ballistic and adjust phases.}
    \label{fig:effectormotion}
\end{figure}

\begin{figure}[ht!]
    \centering
    \includegraphics[width=0.75\linewidth]{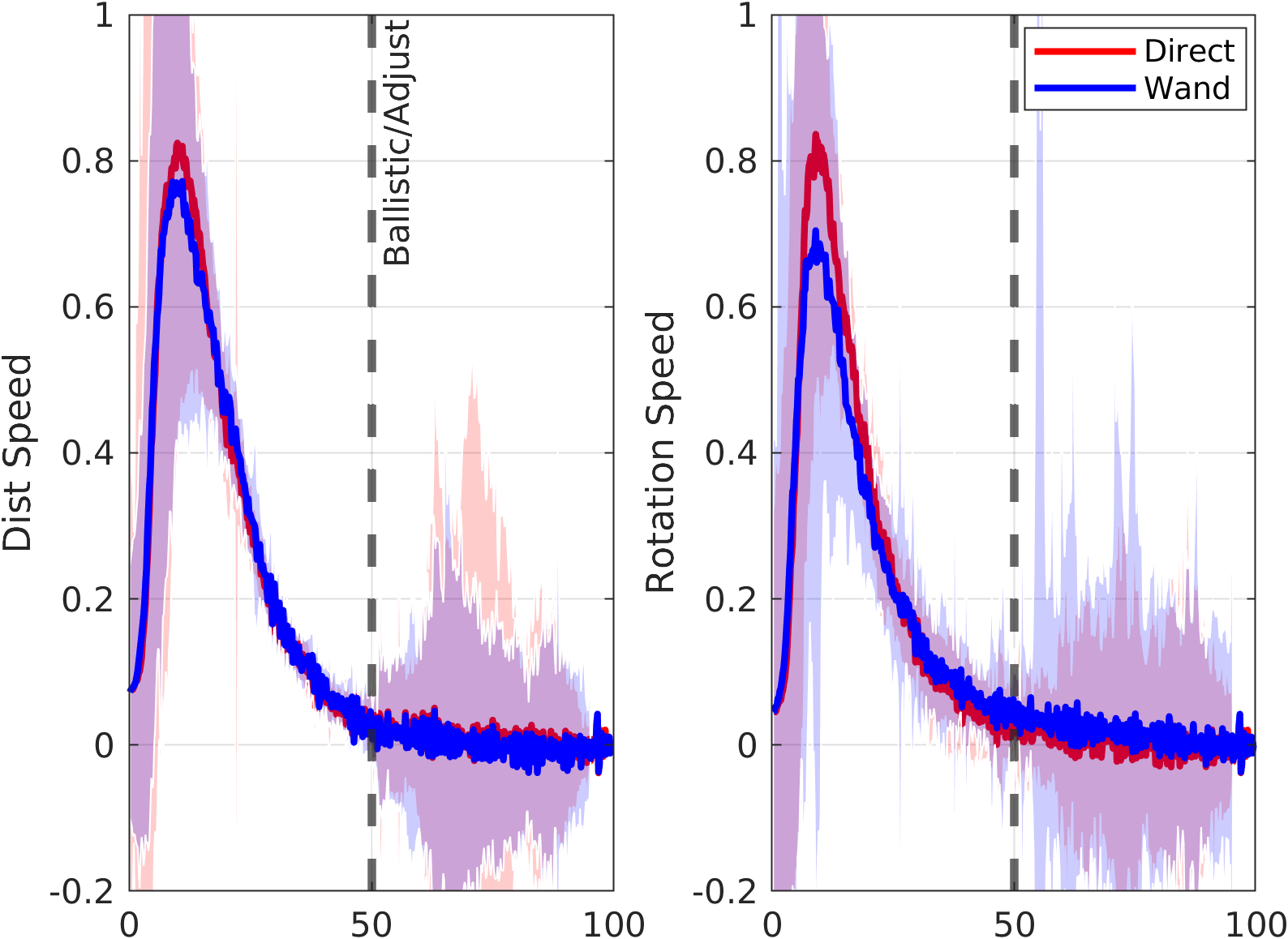}
    \caption{Median of the Hand Velocity in Normalized Translation and Rotation Trajectories. The lighter area denotes the standard deviation of the curve during trials with Visualisation. The dotted line indicates the separation between the estimated ballistic and adjust phases.}
    \label{fig:effectorspeed}
\end{figure}

We also present the desired effector (whether the tip of the virtual wand, or the projected hand motions) normalized motions as well as velocities Fig.\ref{fig:effectormotion} and Fig.\ref{fig:effectorspeed} respectively during trials with Visualisation. Similar curves are observed without visualisation but exhibits higher standard deviations. 

Though the median velocity curve only exhibits one single motion, the standard deviation of the velocity actually shows that the average filter mitigates the presence of a secondary motion to adjust the effector's position. Therefore, we denote that the global motion consist of two sub-motions : a ballistic motions, which, based on the standard deviation of the velocity, accounts for 50\% of the duration, and an adjusting motion accounting for the remaining 50\%. This temporal distribution is consistent across both mappings. We also note that the ballistic motion accounts for 80 up to 95\% of the total motion amplitude, while the adjust motion for the remaining 5 to 20\%.


\subsubsection{Duration of ballistic and adjust phase} 
\begin{figure}[ht!]
    \centering
    \includegraphics[width=0.75\linewidth]{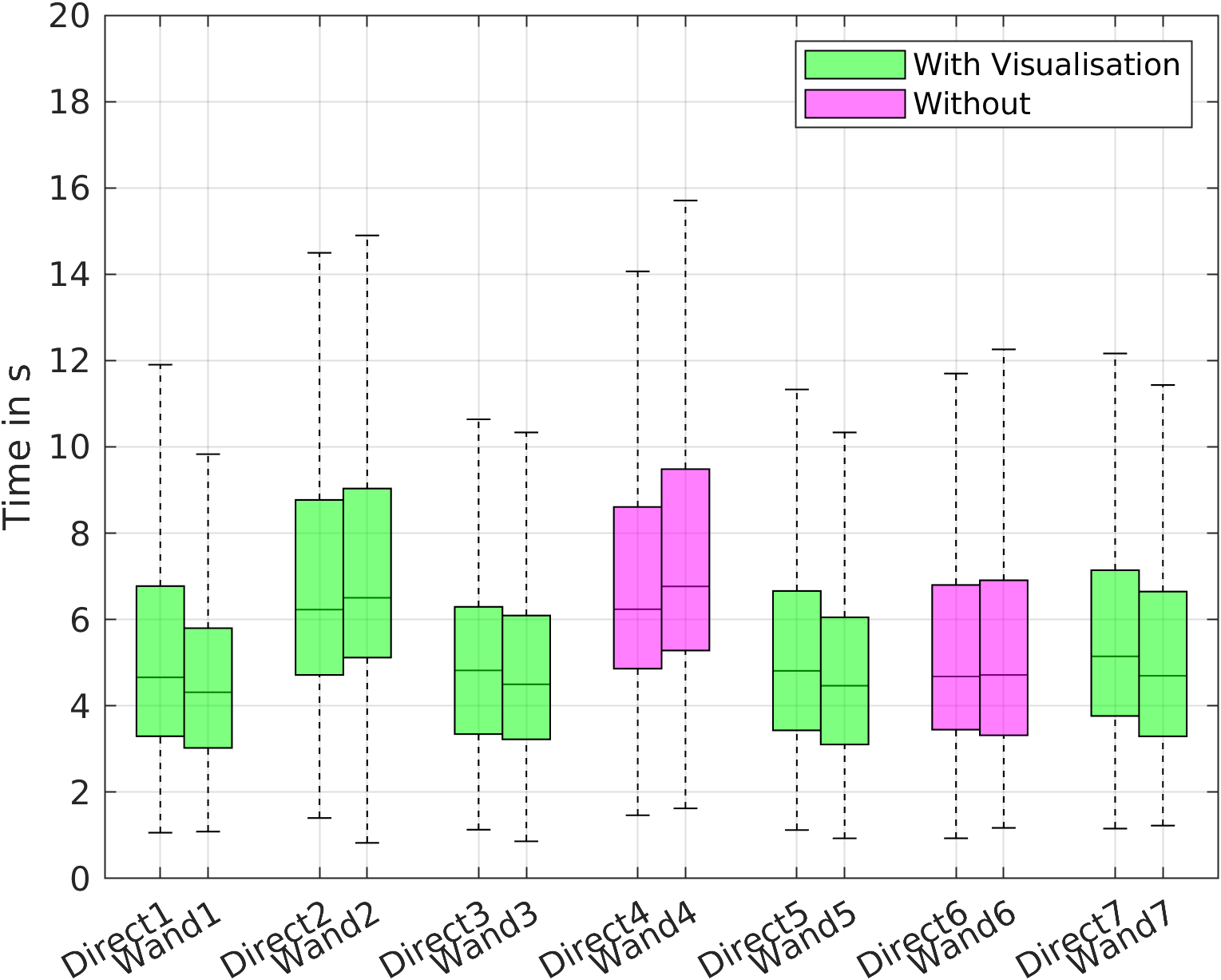}
    \caption{Distribution of the response time of the effector at 80\% per target in s.}
    \label{fig:times95}
\end{figure}

In order to try to evaluate the time of the ballistic motion and the time of the adjust motion, we also present the 80\% response time of the end-effector in Fig.\ref{fig:times95}. The value 80\% was chosen as the lowest value of the standard deviation curve in position and rotation at the end of the ballistic motion Fig.\ref{fig:effectorspeed}. The results are consistent with the overshoot results with more important times during the 2nd and 4th trial, indicating that the ballistic motion was less precise during these phases, which probably caused the overshoot.

\subsection{Analysis of hand motions during ballistic phase}
\begin{figure}[h]
     \centering
     \begin{subfigure}[t]{0.23\textwidth}
         \centering
         \includegraphics[width=\textwidth]{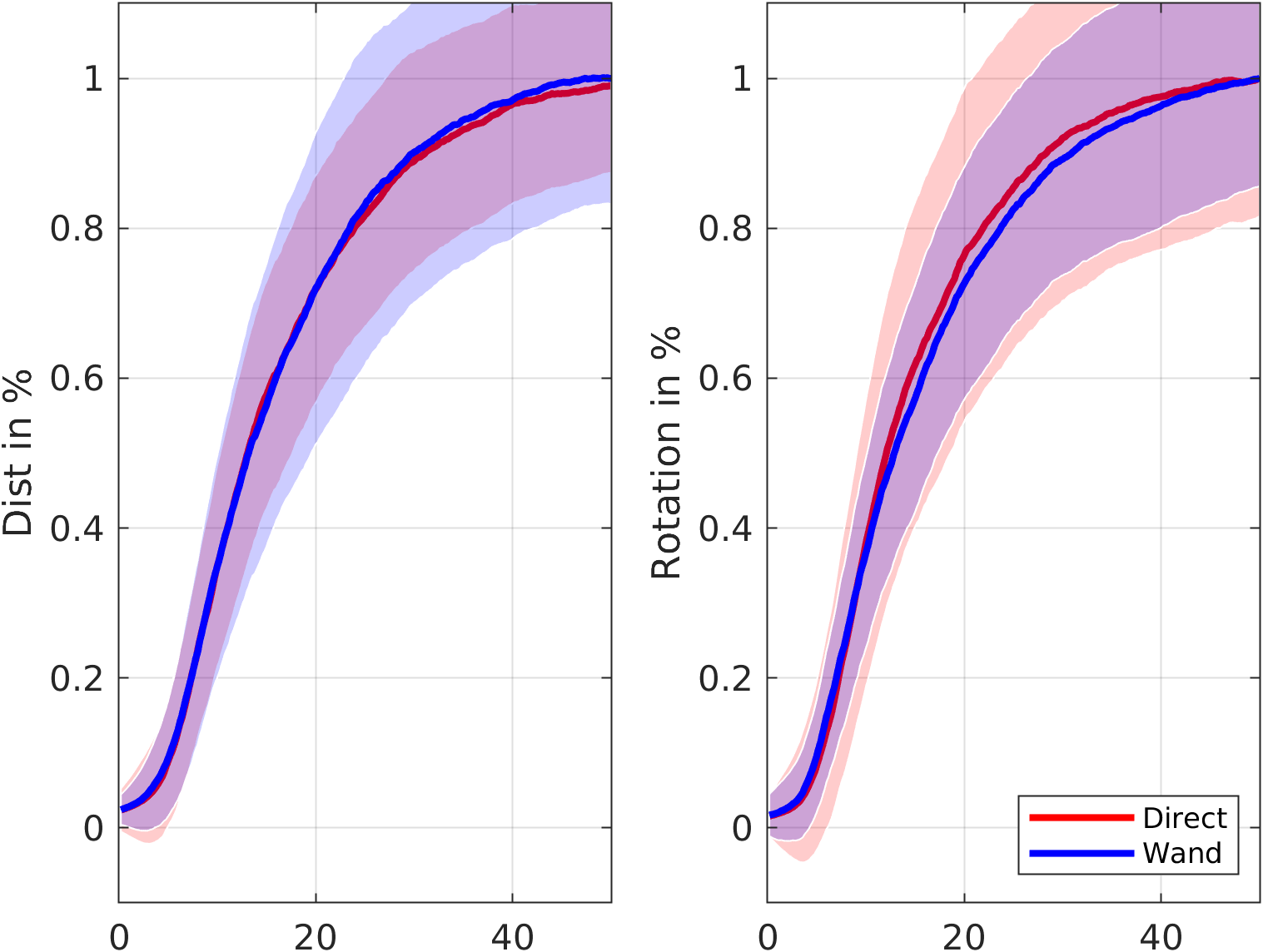}
     \end{subfigure}
     \begin{subfigure}[t]{0.23\textwidth}
         \centering
         \includegraphics[width=\textwidth]{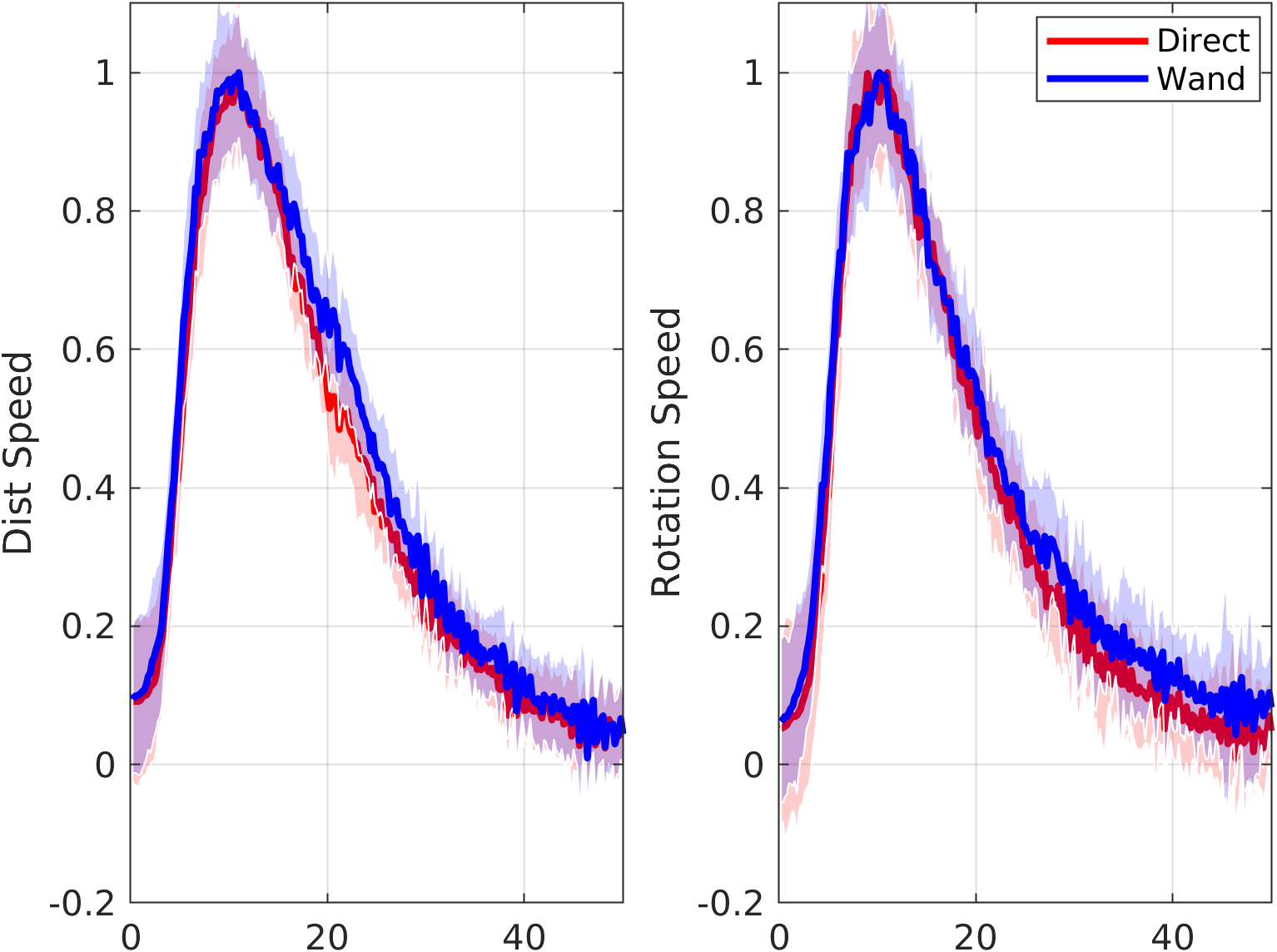}
     \end{subfigure}
     \caption{Normalized Median Hand Trajectories during Ballistic Phase in Translation and Rotation (left) and the associated normalized velocities during Ballistic Phase (right).}
     \label{fig:handspeed}
\end{figure}

We also want to analyze is there are any motor control changes between the two mappings. Though the targets have been designed to that the optimal hand trajectory is the same, we can also compare the hand motions and velocities. As suggested by \cite{torres_simultaneous_2004}, we can analyze motor control invariant kinematic properties during the ballistic motion of the hand. This approach involves normalizing the ballistic motions in both time and amplitude as shown Fig.\ref{fig:handspeed}. The plots show very similar temporal behaviours in both translation and orientation.

\begin{figure}[ht!]
    \centering
    \includegraphics[width=0.75\linewidth]{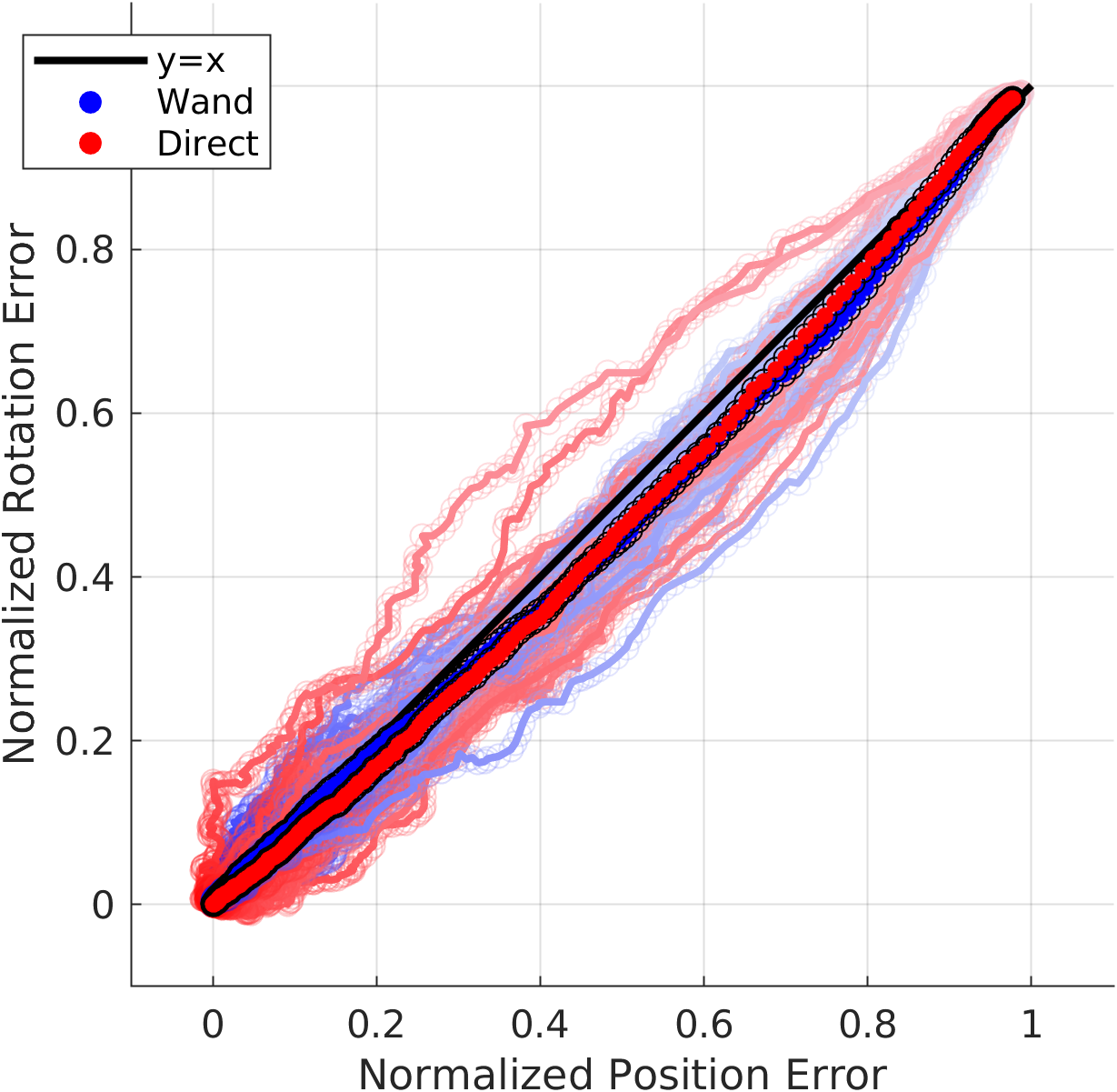}
    \caption{Normalized Hand Spatial Coordination during Ballistic Phase per participants. The darker curves with black outlines indicates the average across all participants (the blue-wand one and the red-direct one are almost superimposed), while the lighter curves are per participant.}
    \label{fig:handcoordinationperparticipants}
\end{figure}

We can also plot the motions curves in translation and orientation from Fig.\ref{fig:handspeed} as a function of one another in order to study the spatial coordination behaviour. According to \cite{torres_simultaneous_2004}, an invariant spatial coordination should be observed, with the rotation error plotted against the position error forming a linear pattern. As these plots are averaged on 20 participants, which might hide participant-specific differences, we also provide the same plot per participants Fig.\ref{fig:handcoordinationperparticipants}. Individual participant analysis revealed that 18 out of 20 participants exhibited very similar patterns between the two modes, with only 2 participants showing patterns above the y=x line in Direct Mapping mode and below the y=x in Wand Mapping mode.

As a comparison, we provide the same spatial coordination plot expressed at the tip of the effector during ballistic motion. The curve in Direct mode remains identical, as the relative motions of the hand and of the effector are one-to-one mapped, but the curve in Wand Mode switches from below the y=x curve for the hand, to above the y=x axis for the effector, as, indeed, the wand mapping mostly transforms rotations into large translations and conversely large translations into rotations of the effector, inverting the coordination pattern between the hand and the tip of the effector. This behavior is not only observed in the collective average but also in the individual behaviors of each participant.

\begin{figure}[ht!]
    \centering
    \includegraphics[width=0.75\linewidth]{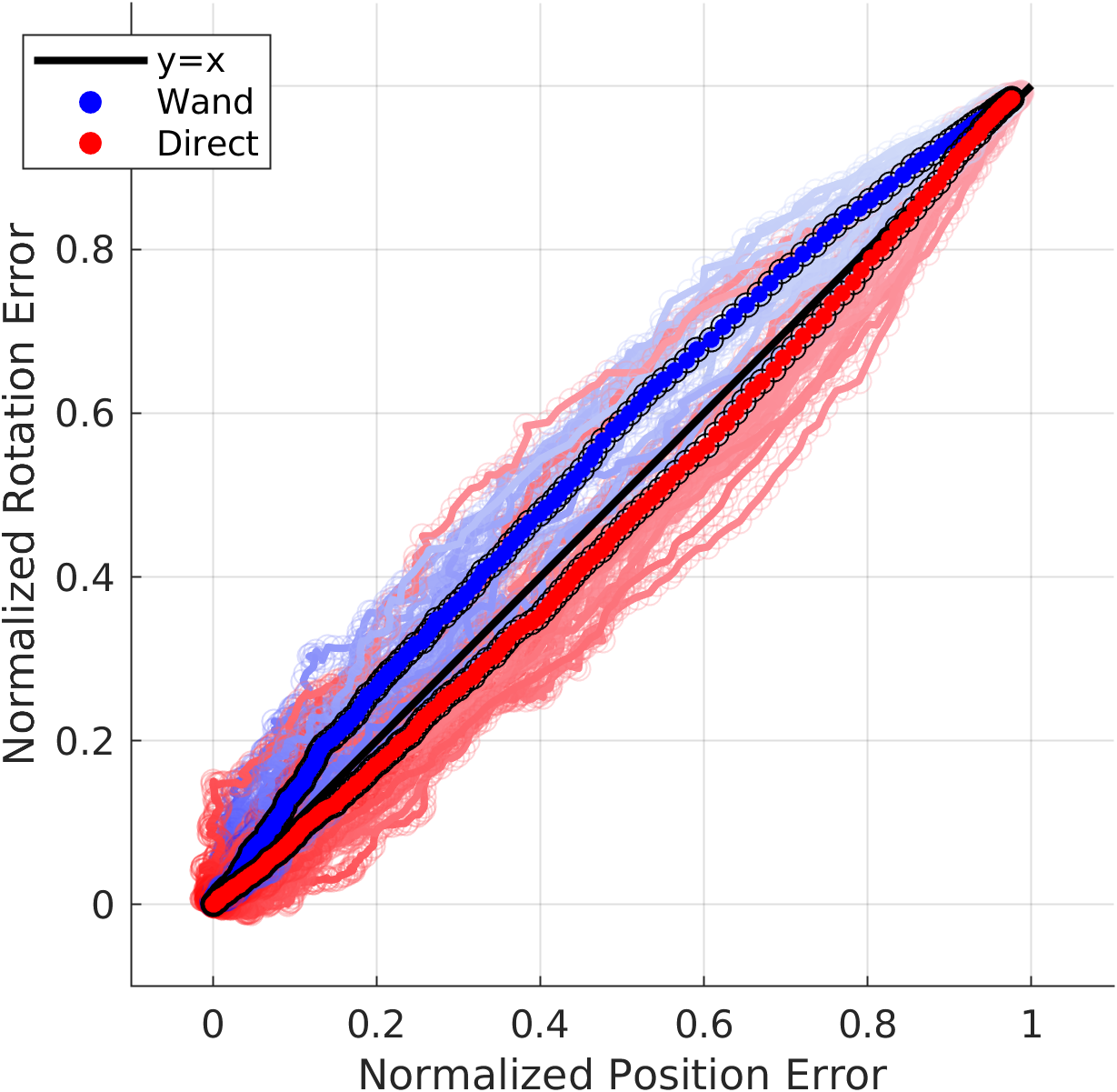}
    \caption{Normalized Desired Effector Spatial Coordination during Ballistic Phase averaged across all participants (darker line with black outline) and per participant (lighter curves).}
    \label{fig:effectorcoordinationperparticipants}
\end{figure}

\subsection{Preferences}
\begin{figure}[ht!]
    \centering
    \includegraphics[width=0.75\linewidth]{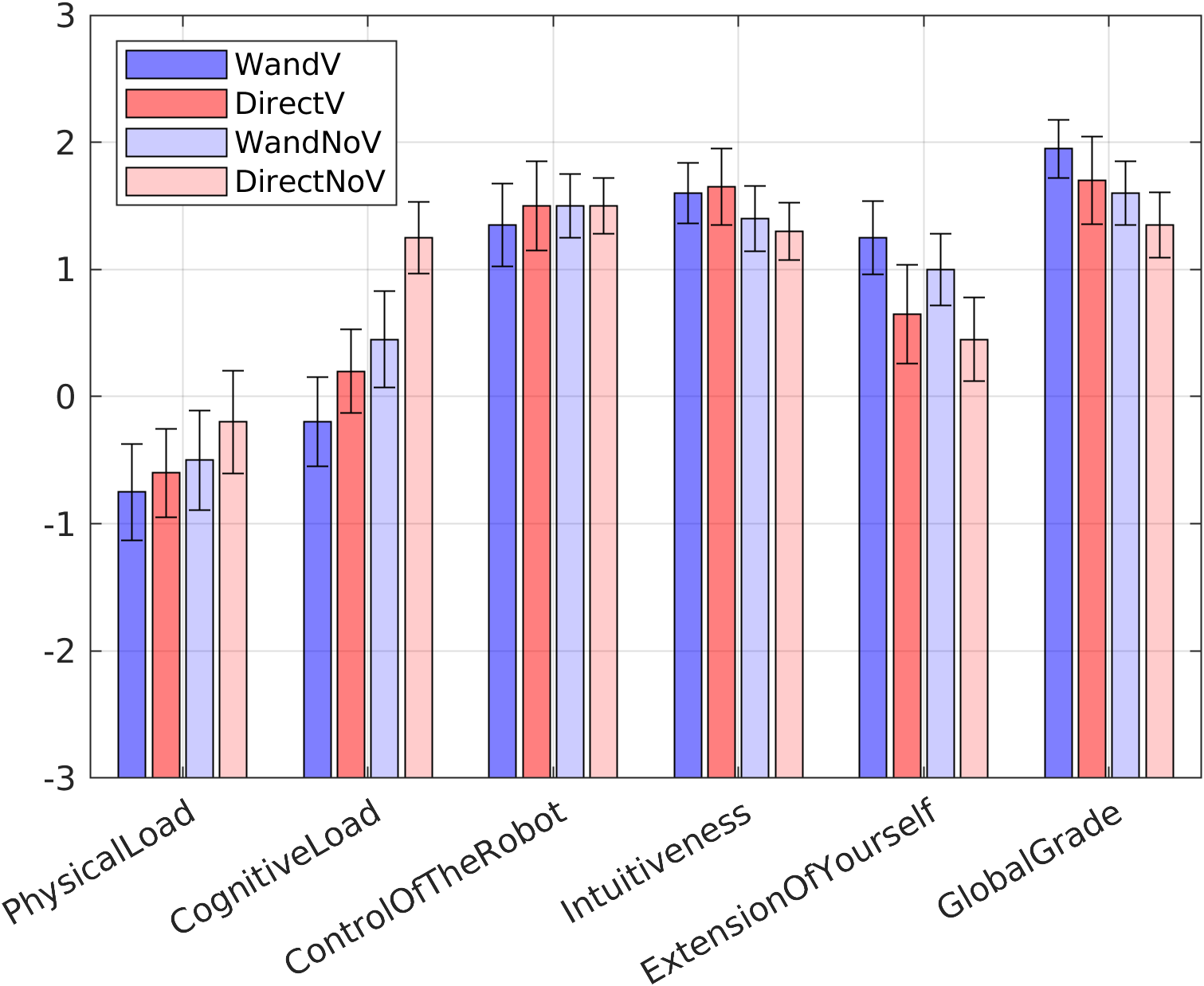}
    \caption{Questionnaires responses. A high rated answer for questions~1 and~2 corresponds to a negative characteristic of the system while for questions~3 to~6 it reflects a positive characteristic.}
    \label{fig:questionnaires}
\end{figure}

\begin{figure}[ht!]
    \centering
    \includegraphics[width=\linewidth]{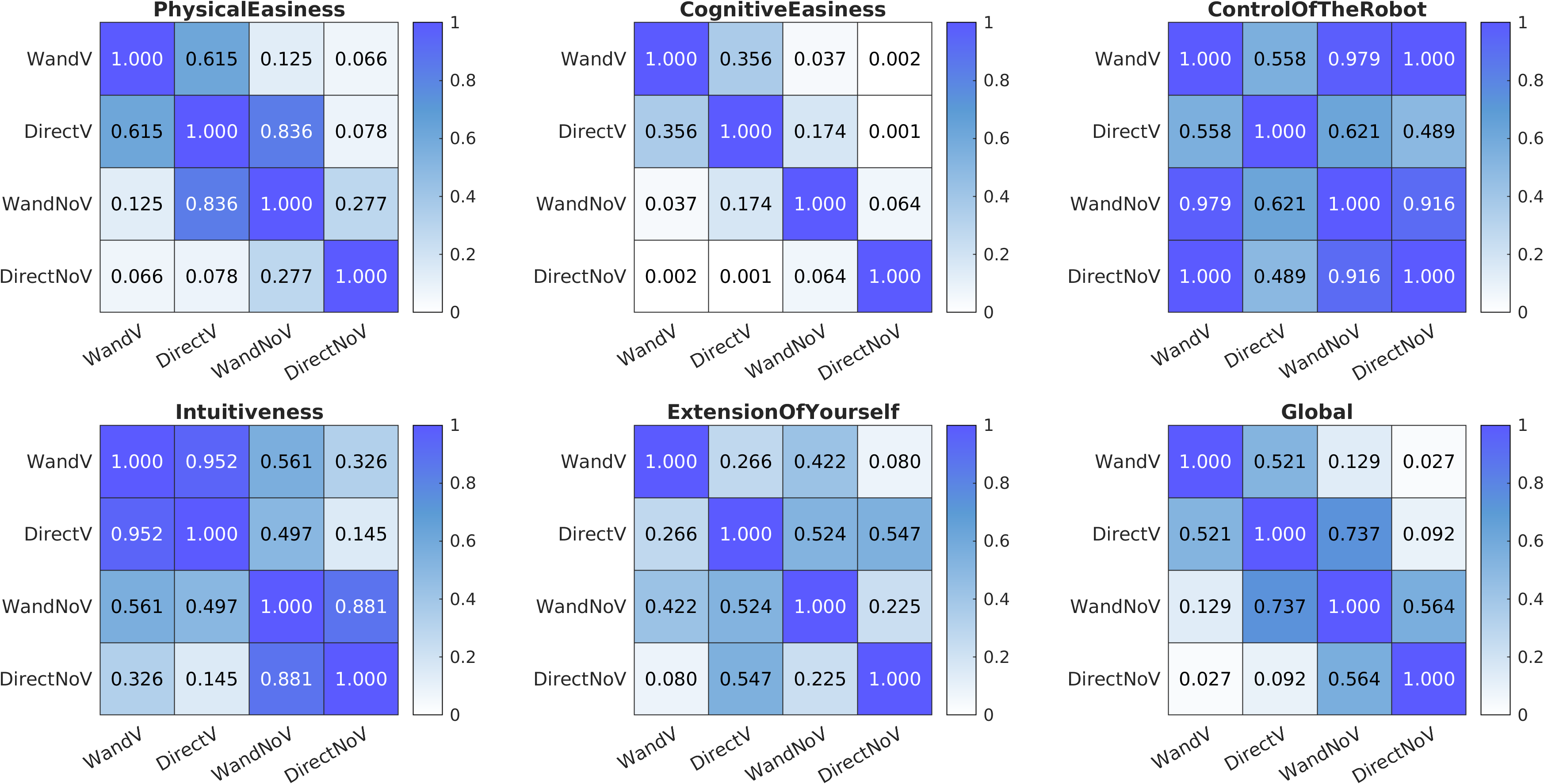}
    \caption{2-by-2 p-Values associated with each mode and each question.}
    \label{fig:questionnairepvalues}
\end{figure}

Finally we provide the questionnaires' results on Fig.\ref{fig:questionnaires}. In order to establish significant statistical difference, the 2-by-2 p values comparison are also provided Fig.\ref{fig:questionnairepvalues}. The removal of visual cues in Direct Mapping significantly increased cognitive load according to the results, but the questionnaire responses remained closely aligned across both mapping modes and were generally positive in all conditions with some non-significative preferences for the Wand Mapping.


\section{Discussion}
The performances between the two mapping modes were found to be nearly identical, with both the durations to complete the tasks and the overshoots exhibiting a strong similarity. Orientation overshoot was slightly higher in the Direct Mapping mode, though not exceeding 10\%. This variance may be attributed to the fact that the pure orientation of the effector requires larger motions in Wand Mapping.

In the 2nd trial, there was an increase in overshoot without a corresponding increase in time. It's plausible that participants, informed that the targets remained the same, attempted to expedite their movements, inadvertently resulting in higher overshoot. Though, the operator being faster than the Kinova robot itself, it did not impacted the overall task duration. Interestingly, this behavior was noted in both mappings. In the 3rd trial, participants in both mappings demonstrated a reduction in overshooting, suggesting an awareness that the overshoot strategy was sub-optimal, despite the delay in the real robot's response.

Upon introducing visual perturbations, both overshoot and completion time increased. In the second perturbed trial, overshoot significantly decreased, accompanied by a reduction of the 80\% response time. This suggests that participants achieved a better integration of the mapping, leading to more accurate ballistic motions. However, the precision and the adjustment phase remained challenging without visualisation, indicating that further training may be necessary to enhance these aspects. Remarkably, performances without visualization were close to one another for both mappings, suggesting that both mappings were similarly integrated in the body schema.

The overall behavior of normalized effector motion was also notably similar between the two modes. However, one can see that in both mode the standard deviation of the speed curve reveals two distinct sub-motions: a ballistic motion accounting for 50\% of the duration and an adjustment motion accounting for the remaining 50\%. The ballistic motion, in turn, represented 80 to 95\% of the overall effector motion in distance and rotation.

Then, to examine potential effects of mapping on motor control, an analysis of hand motions was conducted, focusing on ballistic phase. The median spatial coordination indicated a slight deviation below the y=x curve, possibly attributable to the robot's delay which is slightly more pronounced in rotation due to the arm's geometry and reconfiguration challenges, but is overall close to being linear. Interestingly, again, the median pattern remained highly similar between the two mapping modes while the effector task was different for the two modes. Notably, when comparing the same plot for the end-effector, a clear difference emerged, with the wand mapping consistently following a pattern above the y=x line. This suggests that hand motions and motor control were not significantly affected by the mapping choice whereas the effector was exhibiting different trajectories and behaviours, which supports that constructing targets to elicit identical optimal hand motions for both modes was indeed observed.

In the context of questionnaires, no major differences were observed between Wand and Direct Mapping. This overall consistency in participant feedback supports the conclusion that the mapping did not significantly impact motor control, and participants generally perceived both modes positively for all the listed criteria.

Based on the comprehensive observations and analyses of performance metrics, participant preferences, and motor control dynamics conducted in this study, our findings suggest that the choice between Direct and Wand mapping in a general hand scenario — exploring various translations and orientations - does not appear to be of high significance, and both mapping modes demonstrated comparable outcomes.

However, our knowledge of the mapping prompt us to consider specific scenarios where the choice of mapping may indeed hold significance. In instances where the task necessitates translations with minimal rotations, Wand Mapping could prove advantageous. The leverage provided by the wand, particularly in scenarios involving large translations, may contribute to enhanced control and improved overall performance. Conversely, in situations requiring substantial rotational reconfigurations alongside minor translational displacements, Direct Mapping may be the preferred choice.

Moreover, the suitability of each mapping method might also depend on the specific body part employed for robot control. Wand Mapping, for instance, could offer advantages when using body parts with significant rotational capabilities, such as the head or trunk (as to be found in the case of assistance to paralyzed users), allowing for the scaling of rotations to achieve substantial robotic translations due to the leverage provided by the wand. this scaling not only converts rotations into translations but is exclusively grounded in the length of the link. Therefore, it relies on geometric considerations of the task, differing from traditional translation scaling that necessitates the adjustment of a gain. Oppositely, Direct Mapping holds an advantage to deal with important rotations without translating the effector, which is hardly possible with the Wand Mapping.

In essence, \textbf{the optimal choice between Direct and Wand Mapping should be contingent on the specific requirements of the task, the body part used for control, and the desired balance between rotational and translational movements}. These considerations underscore the importance of tailoring the mapping strategy to the specific characteristics and demands of the robotic manipulation scenario. Likewise, in certain scenarios, such as navigating a wheelchair, position control cannot serve as a substitute for velocity control. Hence, we are contemplating the introduction of hybrid modes that would permit transitioning between Wand, Direct, or Velocity control. Moreover, these hybrid modes could incorporate the simultaneous utilization of multiple controls, such as employing a joystick to manage the length of the wand while employing head orientations to position the desired effector's position.

\section{Conclusion}

In this paper, we introduced an innovative control paradigm, termed "wand" mapping or "rotation-to-position" mapping, which utilizes a virtual linkage mapping for 6 DoF manipulation. The primary advantage of this approach lies in its ability to translate body rotations into substantial effector translations, albeit at the cost of a reduced rotational task space. A comparative analysis with the well-established one-to-one "direct" position mapping, in a general 6 DoF scenario with equivalent hand displacements, reveals comparable performance, user preferences, and motor control behaviors across both mappings.

The objective of this wand mapping is not to supplant the direct mapping method but rather to offer an alternative that may prove more suitable in specific scenarios. For example, the amplification could prove beneficial in scenarios involving body parts characterized by a wide rotational range and limited translation range, such as the head or trunk.

While the amplifying mode offers advantages, the suitability of this mapping depends on the nature of the task and, for applications demanding intricate orientations with fixed translations, an automatic handling approach, such as vision-based grasping, may be needed. A switch between direct and wand mapping could also prove advantageous if operators can seamlessly transition between the two modes.

\hypersetup{urlcolor = black}

\UseRawInputEncoding
\bibliography{BodyInterface.bib}

\begin{thebibliography}{10}
\providecommand{\url}[1]{#1}
\csname url@samestyle\endcsname
\providecommand{\newblock}{\relax}
\providecommand{\bibinfo}[2]{#2}
\providecommand{\BIBentrySTDinterwordspacing}{\spaceskip=0pt\relax}
\providecommand{\BIBentryALTinterwordstretchfactor}{4}
\providecommand{\BIBentryALTinterwordspacing}{\spaceskip=\fontdimen2\font plus
\BIBentryALTinterwordstretchfactor\fontdimen3\font minus
  \fontdimen4\font\relax}
\providecommand{\BIBforeignlanguage}[2]{{%
\expandafter\ifx\csname l@#1\endcsname\relax
\typeout{** WARNING: IEEEtran.bst: No hyphenation pattern has been}%
\typeout{** loaded for the language `#1'. Using the pattern for}%
\typeout{** the default language instead.}%
\else
\language=\csname l@#1\endcsname
\fi
#2}}
\providecommand{\BIBdecl}{\relax}
\BIBdecl

\bibitem{noauthor_world_2023}
``\BIBforeignlanguage{en}{World {Robotics} 2023, {IFR} presentation},'' 2023.

\bibitem{chung_functional_2013}
C.-S. Chung, H.~Wang, and R.~A. Cooper, ``\BIBforeignlanguage{en}{Functional
  assessment and performance evaluation for assistive robotic manipulators:
  {Literature} review},'' \emph{\BIBforeignlanguage{en}{The Journal of Spinal
  Cord Medicine}}, vol.~36, no.~4, 2013.

\bibitem{al-halimi_performing_2016}
R.~K. Al-Halimi and M.~Moussa, ``\BIBforeignlanguage{en}{Performing {Complex}
  {Tasks} by {Users} {With} {Upper}-extremity disabilities {Using} a 6-{DOF}
  {Robotic} {Arm}: {A} {Study}},'' \emph{\BIBforeignlanguage{en}{IEEE
  TRANSACTIONS ON NEURAL SYSTEMS AND REHABILITATION ENGINEERING}}, 2016.

\bibitem{casadio_body-machine_2012}
\BIBentryALTinterwordspacing
M.~Casadio, R.~Ranganathan, and F.~A. Mussa-Ivaldi,
  ``\BIBforeignlanguage{en}{The {Body}-{Machine} {Interface}: {A} {New}
  {Perspective} on an {Old} {Theme}},'' \emph{\BIBforeignlanguage{en}{Journal
  of Motor Behavior}}, vol.~44, no.~6, pp. 419--433, Nov. 2012. [Online].
  Available:
  \url{http://www.tandfonline.com/doi/abs/10.1080/00222895.2012.700968}
\BIBentrySTDinterwordspacing

\bibitem{parietti_supernumerary_nodate}
F.~Parietti and H.~Asada, ``\BIBforeignlanguage{en}{Supernumerary {Robotic}
  {Limbs} for {Aircraft} {Fuselage} {Assembly}: {Body} {Stabilization} and
  {Guidance} by {Bracing}}.''

\bibitem{pulgarin_assessing_nodate}
E.~J.~L. Pulgarin, O.~Tokatli, G.~Burroughes, and G.~Herrmann,
  ``\BIBforeignlanguage{en}{Assessing tele-manipulation systems using task
  performance for glovebox operations},''
  \emph{\BIBforeignlanguage{en}{Frontiers in Robotics and AI}}.

\bibitem{lincoln_visual_1953}
\BIBentryALTinterwordspacing
R.~C. Lincoln, ``\BIBforeignlanguage{en}{Visual tracking: {III}. {The}
  instrumental dimension of motion in relation to tracking accuracy.}''
  \emph{\BIBforeignlanguage{en}{Journal of Applied Psychology}}, vol.~37,
  no.~6, pp. 489--493, Dec. 1953. [Online]. Available:
  \url{http://doi.apa.org/getdoi.cfm?doi=10.1037/h0060608}
\BIBentrySTDinterwordspacing

\bibitem{won_kim_comparison_1987}
\BIBentryALTinterwordspacing
{Won Kim}, F.~Tendick, S.~Ellis, and L.~Stark, ``\BIBforeignlanguage{en}{A
  comparison of position and rate control for telemanipulations with
  consideration of manipulator system dynamics},''
  \emph{\BIBforeignlanguage{en}{IEEE Journal on Robotics and Automation}},
  vol.~3, no.~5, pp. 426--436, Oct. 1987. [Online]. Available:
  \url{http://ieeexplore.ieee.org/document/1087117/}
\BIBentrySTDinterwordspacing

\bibitem{noauthor_virtuose_nodate}
\BIBentryALTinterwordspacing
``Virtuose™ {6D} {RV} - {HAPTION} {SA}.'' [Online]. Available:
  \url{https://www.haption.com/fr/products-fr/virtuose-6d-fr.html}
\BIBentrySTDinterwordspacing

\bibitem{dominjon_comparison_nodate}
L.~Dominjon, A.~Lécuyer, J.-M. Burkhardt, and R.~Simon,
  ``\BIBforeignlanguage{en}{A {Comparison} of {Three} {Techniques} to
  {Interact} in {Large} {Virtual} {Environments} {Using} {Haptic} {Devices}
  with {Limited} {Workspace}}.''

\bibitem{freschi_technical_2013}
\BIBentryALTinterwordspacing
C.~Freschi, V.~Ferrari, F.~Melfi, M.~Ferrari, F.~Mosca, and A.~Cuschieri,
  ``\BIBforeignlanguage{en}{Technical review of the da {Vinci} surgical
  telemanipulator: {Technical} review of the da {Vinci} surgical
  telemanipulator},'' \emph{\BIBforeignlanguage{en}{The International Journal
  of Medical Robotics and Computer Assisted Surgery}}, vol.~9, no.~4, pp.
  396--406, Dec. 2013. [Online]. Available:
  \url{https://onlinelibrary.wiley.com/doi/10.1002/rcs.1468}
\BIBentrySTDinterwordspacing

\bibitem{jewell_custom-made_1989}
K.~H. Jewell, ``A custom-made head pointer for children,'' \emph{The American
  Journal of Occupational Therapy}, vol.~43, no.~7, pp. 456--460, 1989,
  publisher: The American Occupational Therapy Association, Inc.

\bibitem{dymond_controlling_1996}
E.~Dymond and R.~Potter, ``Controlling assistive technology with head
  movements-a review,'' \emph{Clinical rehabilitation}, vol.~10, no.~2, pp.
  93--103, 1996, publisher: Sage Publications Sage CA: Thousand Oaks, CA.

\bibitem{maravita_tools_2004}
\BIBentryALTinterwordspacing
A.~Maravita and A.~Iriki, ``\BIBforeignlanguage{en}{Tools for the body
  (schema)},'' \emph{\BIBforeignlanguage{en}{Trends in Cognitive Sciences}},
  vol.~8, no.~2, pp. 79--86, Feb. 2004. [Online]. Available:
  \url{https://linkinghub.elsevier.com/retrieve/pii/S1364661303003450}
\BIBentrySTDinterwordspacing

\bibitem{nuzzi_hands-free_2020}
C.~Nuzzi, S.~Ghidini, R.~Pagani, S.~Pasinetti, G.~Coffetti, and G.~Sansoni,
  ``\BIBforeignlanguage{en}{Hands-{Free}: a robot augmented reality
  teleoperation system},'' \emph{\BIBforeignlanguage{en}{International
  Conference on Ubiquitous Robots}}, 2020.

\bibitem{gonzalez_measurement_2010}
\BIBentryALTinterwordspacing
Ã.~González, ``\BIBforeignlanguage{en}{Measurement of areas on a sphere using
  {Fibonacci} and latitude-longitude lattices},''
  \emph{\BIBforeignlanguage{en}{Mathematical Geosciences}}, vol.~42, no.~1, pp.
  49--64, Jan. 2010, arXiv:0912.4540 [math]. [Online]. Available:
  \url{http://arxiv.org/abs/0912.4540}
\BIBentrySTDinterwordspacing

\bibitem{torres_simultaneous_2004}
\BIBentryALTinterwordspacing
E.~B. Torres and D.~Zipser, ``\BIBforeignlanguage{en}{Simultaneous control of
  hand displacements and rotations in orientation-matching experiments},''
  \emph{\BIBforeignlanguage{en}{Journal of Applied Physiology}}, vol.~96,
  no.~5, pp. 1978--1987, May 2004. [Online]. Available:
  \url{https://www.physiology.org/doi/10.1152/japplphysiol.00872.2003}
\BIBentrySTDinterwordspacing

\end{thebibliography}
\bibliographystyle{IEEEtran}

\end{document}